\documentclass[runningheads]{llncs}
\usepackage{graphicx}
\usepackage{csquotes}
\usepackage{algpseudocode}
\usepackage{algorithm}
\usepackage{commath}
\usepackage{xcolor}

\usepackage[english]{babel}

\usepackage{listings}
\usepackage{longtable}
\usepackage{array}
\begin{document}
\title{Using adversarial images to improve outcomes of federated learning for non-IID data}
\titlerunning{Adversarial images in federated learning for non-IID data}
\author{Anastasiya Danilenka\inst{1}\orcidID{0000-0002-3080-0303}
\and
Maria Ganzha\inst{1}\orcidID{0000-0001-7714-4844} \and
Marcin Paprzycki\inst{2}\orcidID{0000-0002-8069-2152}
\and Jacek Mańdziuk\inst{1}\orcidID{0000-0003-0947-028X}}
\authorrunning{A. Danilenka, et al.}
\institute{Faculty of Mathematics and Information Science, Warsaw University of Technology, Koszykowa 75, 00-662 Warszawa, Poland\\
\email{\{anastasiya.danilenka.stud,maria.ganzha,jacek.mandziuk\}@pw.edu.pl}
\and
Systems Research Institute Polish Academy of Sciences\\
\email{marcin.paprzycki@ibspan.waw.pl}}
\maketitle              % typeset the header of the contribution
\begin{abstract}
One of the important problems in \textit{federated learning} is how to deal with unbalanced data. This contribution introduces a novel technique designed to deal with label skewed non-IID data, using adversarial inputs, created by the I-FGSM method. Adversarial inputs guide the training process and allow the Weighted Federated Averaging to give more importance to clients with ``selected'' local label distributions. Experimental results, gathered from image classification tasks, for MNIST and CIFAR-10 datasets, are reported and analyzed.

\end{abstract}
\section{Introduction}

The concept of federated learning (FL) was introduced in~\cite{mcmahan2017communicationefficient}. It aims at creating a shared/global model, combining information from multiple computing nodes, without sharing individual source datasets. In FL, training proceeds in rounds. In each round, clients receive the current version of the global model, perform training using local data, and return the updated model parameters. Thus, the clients in FL pipelines act as both data sources and computational nodes, while the learning process is controlled, and orchestrated, by the server/manager (for needed details see~\cite{mcmahan2017communicationefficient,DBLP:journals/corr/abs-1912-04977}).

The process of combining updates, received from clients, into the new version of the shared model, is a subject of intensive research. However, its most straightforward realization averages all model parameters, returned by all participating clients. This algorithm (FedAvg) has been summarized as Algorithm~\ref{alg:cap} 
(see~\cite{fed_proxy} for additional details).
\begin{algorithm}
\caption{Federated Averaging (FedAvg) algorithm. The number of server-side epochs is $N$, the number of clients participating in the training round is $K$, $m$ are parameters of the model, $s$ is a set of parameters needed for a client to train the model (learning rate, number of local epochs, batch-size,  model-dependent parameters)}\label{alg:cap}
\begin{algorithmic}
\Ensure global model initialization
\Ensure clients ready for training
\For{$i \neq N$}
    \For{ $client_k$ in $clients[1..K]$ in parallel}
        \State $m_{i}^{client_{k}}  \gets RequestModelUpdate(m_{i}, s_i)$
    \EndFor 
    \State $m_{i} \gets {\frac{1}{K}}\times \sum_{k=0}^{K} m_{i}^{client_{k}}$
\EndFor
\end{algorithmic}
\end{algorithm}
A natural extension to FedAvg is weighting, i.e. assigning individual importance to each client, based on some additional measure/knowledge. For example, clients with more local data can be treated as delivering more valuable input to the training process, and thus receive larger weights for updates they ``suggest''~\cite{mcmahan2017communicationefficient}. 

Federated learning pipeline utilizes client devices' local datasets, while the complete dataset is never gathered ``in one place''. Therefore, statistical properties of both, the local datasets and of the global dataset, are unknown. As a result, data may turn out to be not identically independently distributed (non-IID). It has been shown that non-IID datasets can (very) negatively affect the quality of a global model produced by the FL~\cite{fed_learning_non_iid,non_iid_survey,xiao2021experimental}. 

From ``statistical point of view'', non-IID data can be divided into categories, based on the source of heterogeneity~\cite{non_iid_survey}, i.e.: (1) data quantity skew (local datasets differ in size), (2) label distribution skew (different devices have different subsets of labels inside local dataset or have data relying on preference), (3) attribute skew (local data has unique characteristic features, noise, perturbations, etc.), (4) temporal skew (local data distributions differ over time or data was collected in different time periods). Obviously, combinations of skews can materialize in local datasets.

This work concentrates on non-IID datasets with label distribution skew. This type of skew naturally occurs in (image) classification tasks. For instance, individual devices may or may not have the full set of labels in their local datasets. Depending on how many labels are represented inside the local dataset, the skew can be extreme with only $1$ label per local dataset, or a ($C-1$)-label skew, where $C$ is the number of labels in classification task, thus, assuming that each local dataset lacks data for one of the labels.

In this context, a novel method, to overcome the problems caused by label distribution skew in non-IID data, called \textit{Adversarial Federated Learning} (AdFL), is proposed. It is conceptually based on adversarial attacks and is applied to neural network based training with image-based data. Thus, we proceed as follows. In Section~\ref{adv_theory}, we introduce key concepts of adversarial attacks. We follow, in Section~\ref{alg-desc}, with the description of the AdFL method. Section~\ref{experimental} presents the experimental setup and results of experiments, conducted for MNIST~\cite{lecun-mnisthandwrittendigit-2010} and CIFAR-10~\cite{cifar_ref} datasets. Conclusions and directions for future research are covered in the last section. 

It should be noted that, due to the space limitation, and the fact that a number of fresh and comprehensive literature reviews already exist for federated learning (see, for instance,~\cite{non_iid_survey,DBLP:journals/corr/abs-1912-04977,LI2020106854}), we omit the standard state-of-the-art discussion. Only references directly needed, in the context of this contribution, are cited.

\section{Adversarial inputs} \label{adv_theory}

In 2014 it was shown that a deep neural network can be misled and inclined to make mistakes when classifying slightly modified images that still remain very easy to categorize by humans. Moreover, these images have been initially correctly classified by the model itself~\cite{https://doi.org/10.48550/arxiv.1312.6199}. Specifically, it was shown that by altering the original input image with small non-random perturbations, one can create an ``adversarial image'' that is indistinguishable from the original one by humans, but forces a classifier to make a mistake.

Overall, adversarial attacks can be divided into two categories: targeted (an initial image is modified in order to be misclassified as a certain class), and untargeted (modifications are made to mislead the model and misclassify the image as ``any'' incorrect class)~\cite{Li_2022}. 

The source image can be modified using various techniques. For example, methods belonging to the family of gradient-based techniques, like Fast Gradient Sign Method (FGSM), or Projected Gradient Method (PGM)~\cite{adv_example_explained,projected_gradients} can be used. These methods rely on having an access to model's gradients. FGSM can be described using the following formula (cf.~\cite{adv_example_explained} for more details):
\begin{center}
   $\eta = \epsilon sign(\nabla_x L(\theta, x, y))$ 
\end{center}
Here, $x$ stands for an initial input,  $y$ -- for the input's label (for tasks that assume labels), $\theta$ -- model's parameters, $L(\theta, x, y)$ -- a loss function (for example, cross entropy for multi-class classification task), $\nabla_x$ -- a gradient computed with back propagation, $\epsilon$ -- a constant chosen to control the amount of perturbations added to the input.

Moreover, the property of transferability of adversarial inputs was discovered~\cite{https://doi.org/10.48550/arxiv.1312.6199}. Specifically, it has been established that adversarial inputs generated by one model can still mislead another similar model, even if their training subsets are non-overlapping. Conducted studies showed that adversarial inputs are most likely derived from the adversarial subspaces~\cite{space_transfer}. A number of contributions has been devoted to discovering ways of improving transferability, understanding it's nature and boundaries~\cite{study_transf,https://doi.org/10.48550/arxiv.2108.07033,https://doi.org/10.48550/arxiv.1907.10823}. The transferability property of adversarial images also provides a possibility for decision-based attacks. Here, a local model is trained on a data subset and tries to learn how to mimic the decision boundary of the model that is a target of the attack~\cite{transferable_attack}. Overall, the success rate for adversarial attacks, depending on various conditions, is in the range from 63\% to 100\%, including cross-model attacks scenarios (where the source model and the target model have different architectures)~\cite{demystif_transfer}.

In what follows, features of adversarial attacks, i.e.: (1) transferability, and (2) connectivity with decision boundary of the model, will be used in the context of FL pipeline to deal with label distribution skew non-IID data.

\section{Description of the proposed AdFL algorithm}\label{alg-desc}

Overall, AdFL utilizes transferability of adversarial inputs, and uses them to improve quality of FL. Here, adversarial inputs are created not as threats to misguide models into making mistakes, but are instantiated to measure the ``cohesion'' among federated clients. This allows to modify the training process without sharing local data. 

As stated in Section~\ref{adv_theory}, for gradient based attacks, having only an access to the model parameters would be sufficient to successfully perform a targeted adversarial attack. Interestingly, an access to the model parameters is exactly what the server in the federated learning pipeline is granted, while knowing nothing about the local (client) data. In particular, it is possible for the server to use updates of individual model parameters, returned by clients, to produce an adversarial image for every label that is defined in given classification task. Adversarial images, in this case, serve as a ``communication tool'' for updated models to check if the produced adversarial samples are able to fit inside decision boundaries of the updated models. 

\begin{figure} [htbp]
    \centering
    \includegraphics[scale=.8]{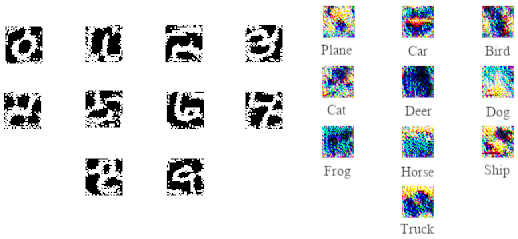}
    \caption{Adversarial images produced for MNIST and CIFAR-10 datasets starting from a totally black image.}
    \label{adv_example}
\end{figure}

Without access to the source data, images consisting of noise, or fully black/white images, are used as the starting point for the adversarial input generation. Examples of images produced from a totally black starting image for MNIST and CIFAR-10 datasets are shown in Figure~\ref{adv_example}.

Note that, the adversarial image generation step occurs on the server side, i.e. for the client devices no changes to the default FL pipeline are needed (no additional information or communication is required). The clients operate ``as always'', i.e. they participate in the training of the global model, using their local data, and send the resulting update proposals back to the server. The enhanced algorithm only changes the way the server behaves, after model updates are gathered from the client devices. Specifically, the server behaviour can be summarized as follows:
\begin{figure} [htbp]
    \centering
    \includegraphics[scale=0.35]{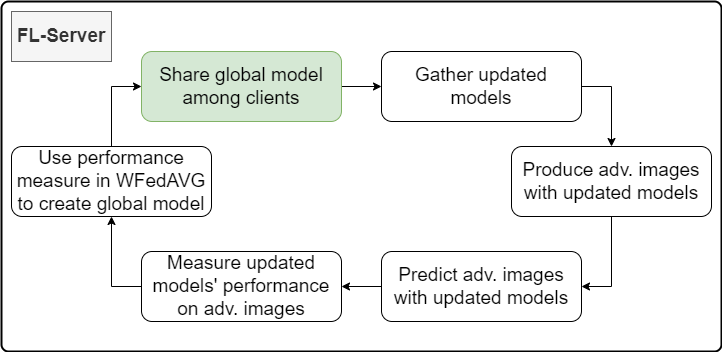}
    \caption{Server-side training epoch routine in AdFL. Green block indicates starting point of each epoch. Validation and testing routines are omitted.}
    \label{flow_diagram}
\end{figure}
\begin{itemize}
  \item Based on obtained parameter updates, the server recreates local models for each client. 
  
  \item Obtained models are used to generate images for each of the labels, according to the iterative version of the Fast Gradient Signed Method (FGSM). Starting from a random noise image, it iteratively modifies the image, using gradients retrieved from the client's model, until the specified number of iterations is reached. 
  
  \item Individual models make predictions of all adversarial inputs generated by all models, including the ones generated ``for itself''. Therefore, each client's model makes $K$ * $C$ predictions, where $K$ -- is the number of clients participating in a given training round and $C$ -- is the number of labels involved in given classification task.
  
 \item Based on the results from the previous step, each client receives a weight that serves as a measure of its ability to predict adversarial inputs produced by other models, as well as ability to produce such images. The evaluation process, used to calculate weigth, is described in Section~\ref{cross_pred}.
  
  \item A weighted version of the  FedAvg algorithm (WFedAvg) is used to generate a global model from the local models, where weighting takes place according to the values calculated during the previous step.
\end{itemize}

The server-side algorithm is summarized in Algorithm~\ref{alg:server}. For clarity, the detailed descriptions of adversarial images generation, and cross-prediction, are omitted, leaving only the high-level representation of steps that are introduced by the proposed algorithm. The diagram featured in Figure~\ref{flow_diagram} shows how each server-side epoch is designed in AdFL.

\begin{algorithm}[htpb]
\caption{
AdFL algorithm. The number of server-side epochs is $N$, the number of clients participating in the training round is $K$, $m$ are parameters of the model, $s$ is a set of parameters needed for a client to train the model (learning rate, number of local epochs, batch-size,  model-dependent parameters), $w$ are weights of the updated versions of global model, $C$ is the number of labels in the considered task}\label{alg:server}
\begin{algorithmic}
\Ensure global model initialization
\Ensure clients ready for training
\For{$i \neq N$}
    \For{ $client_k$ in $clients[1..K]$ in parallel}
        \State $m_{i}^{client_{k}}  \gets RequestModelUpdate(m_{i}, s_i)$
    \EndFor
    \Comment{Produce adversarial images and use them for prediction}
    \For{ $client_k$ in $clients[1..K]$ in parallel}
        \For{ $label_{c}$ in $labels[1..C]$ in parallel}
            \State $images_{label_{c}}^{client_{k}}  \gets CreateAdversarialImage(m_{i}^{client_{k}}, label_{c})$ 
            \For{ $client_j$ in $clients[1..K]$ in parallel}
            \State $predictions_{label_{c}}^{client_{j}}  \gets PredictAdversarialImage(m_{i}^{client_{j}}, images_{label_{c}}^{client_{k}})$ 
            
            \EndFor 
        \EndFor 
    \EndFor 
    
    \Comment{Measure each model performance and use it for WFedAvg}
    
    \State $w \gets DefineModelsWeights(predictions)$  
    
    \State $m_{i} \gets \sum_{k=0}^{K} m_{i}^{client_{k}} \times w_{k}$
\EndFor
\end{algorithmic}
\end{algorithm}

\subsection{Weights calculation} \label{cross_pred}

As stated above, WFedAvg is to be used on the basis of predictions of adversarial images for all classes and for all clients. However, it should be noted that in this process two aspects of image recognition materialize: (a) How well a given client recognizes adversarial images generated by other models (clients)? and (b) How well adversarial images that are produced by a given model (client) are recognized by the other models (clients)? Therefore, the weight assigned to updates originating from a given client will consist of two parts, corresponding to these two aspects.

\subsubsection{The ability of the updated model to predict adversarial images produced by other models} is measured according to Equation (\ref{eq:1}), where each multiplication consists of a binary flag indicating whether or not the prediction for label $c$ generated by model $k$ was correct and the probability returned by the model. The results obtained for the model predicting its own adversarial inputs are omitted.
\begin{equation}
\begin{split}
 \mbox{predicted others} =  \sum_{k=1}^{K}{\sum_{c=0}^{C} \mbox{prediction correct$_{k, c}$} * {\mbox{returned probability$_{k, c}$}}} 
\end{split}
\label{eq:1}
\end{equation}

\subsubsection{The ability of the updated model to produce adversarial images that are recognized by the other models} is defined by Equation (\ref{eq:2}). Again, results for the model ``predicting itself'' are omitted. The formula consists of the binary flag, indicating whether the current model input for label $c$ was correctly predicted by model $k$, and the probability assigned by model $k$ while predicting label $c$. 
\begin{equation}
\begin{split}
\mbox{was predicted} = \sum_{k=1}^{K}{\sum_{c=0}^{C} \mbox{prediction correct$_{k, c}$} * {\mbox{returned probability$_{k, c}$}}} 
\end{split}
\label{eq:2}
\end{equation}

\subsubsection{The total weight} is a result of a simple summation of the scores from Equation (\ref{eq:1}) and Equation (\ref{eq:2}). So, the absolute weight is calculated as:
\begin{equation}
\begin{split}
\mbox{weight total} = \mbox{predicted others} + \mbox{was predicted} 
\end{split}
\label{eq:3}
\end{equation}
After the total weights for all clients that participated in the training epoch, are calculated they are normalized. The final weights for the current epoch are used by the WFedAvg algorithm.

\section{Experimental setup and results}\label{experimental}

\subsection{Experimental setup}

Initial set of experiments has been conducted using two standard benchmark datasets -- MNIST~\cite{lecun-mnisthandwrittendigit-2010} and CIFAR-10~\cite{cifar_ref}. The main interest lied in exploring the performance of the proposed AdFL method, compared to the default FL (with no modifications to the server-side routine). The experiments structure was designed as follows:
\begin{itemize}
    \item In each server-side epoch 5 clients were picked from the training pool.
    \item Clients performed training for 5 epochs and returned results back to the server.
    \item A hold-out IID test dataset was stored on the server, and used to track IID accuracy and per-class accuracy for the global model. The test set was composed of samples that are not presented in any clients' local datasets.
    \item Validation clients were used to modify the learning rate during training. Tracking average validation clients loss was done to adjust the learning rate during the training process, by reducing the learning rate in inverse square root of the number of epochs passed manner.
\end{itemize}
Note that the aim of experiments was \textit{not} to find the set of best ``hyperparameters'' and the most effective ``experimental setup''. Instead, the focus was on proving the potential of AdFL and the premises for its further exploration. This was also the reason for using classic benchmark datasets, which are ``well understood'' and simple enough to focus on the effects of the proposed approach, rather than having them ``hidden'' due to the complexty of the dataset itself.
\begin{table}[htbp]
\caption{Label probability during experiment}
\label{table_prob}
\begin{center}
\begin{tabular}{ | m{6em} || m{0.7cm}| m{0.7cm} | m{0.7cm}|   m{0.7cm}| m{0.7cm}|  m{0.7cm}| m{0.7cm} | m{0.7cm}|  m{0.7cm}| m{0.7cm}|} 
  \hline
   Label & 0 & 1  & 2 & 3 & 4 & 5 & 6 & 7 & 8 & 9 \\ 
  \hline
  \hline
  Probability \% & 3.5 & 4.5 & 10 & 21 & 21 & 20 & 10 & 4.5 & 3.5 & 2 \\ 
  \hline
  \hline
\end{tabular}
\end{center}
\end{table}

Each experiment was conducted multiple times (from 10 to 15 runs), with different model initialization and random seeds. The results were averaged in terms of performance. In all cases, the average standard deviation per label, at the end of training process, was of order of 0.5\%. 

The datasets, generated for the experiments, featured non-IID label skew that was introduced to the local datasets by different label occurrence probabilities. Example of such probability set is given in Table~\ref{table_prob}. The probability was used to generate a subset of labels that were presented on the ``local device''. During experiments with MNIST, only 3 labels were presented in individual local datasets, while for CIFAR-10 experiments the number of unique labels was~4.

\subsection{Experimental results and their analysis}

\begin{figure} [htbp]
    \centering
    \includegraphics[scale=0.45]{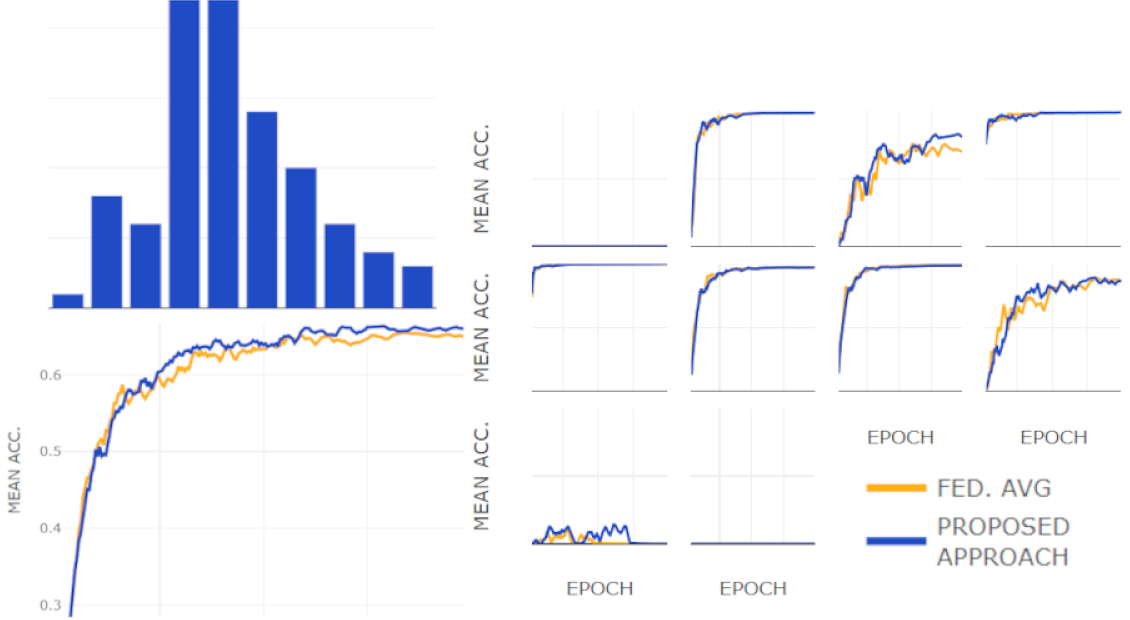}
    \caption{Results of the first MNIST label skew experiment.}
    \label{pop_mnist_label}
\end{figure}

In the first MNIST experiment, the label distribution histogram, illustrating how many clients had the specific label inside their local dataset, the IID test accuracy reported from the holdout dataset, and per class accuracy for the holdout IID dataset, were calculated. The results are shown in Figure~\ref{pop_mnist_label}. The experiment was conducted with several labels being overrepresented (more than 50\% of clients had these labels in local data), these are labels 3, 4, and 5. Labels 1, 2, 6, and 7 were mildly underrepresented (15 to 30\% of client devices had them), and labels 0, 8, and 9 were highly underrepresented.

The total IID accuracy difference between the AdFL and the baseline FL is around 1.5\%. Although a group of labels (0, 1, 4 - 7, and 9) is predicted with the same accuracy by both algorithms, two labels have notable difference in performance: the 3rd label predicted more acurately by 8-11\%, and the gain in accuracy for the 8th label is also around 5-10\%. 

\begin{figure} [htbp]
    \centering
    \includegraphics[scale=0.45]{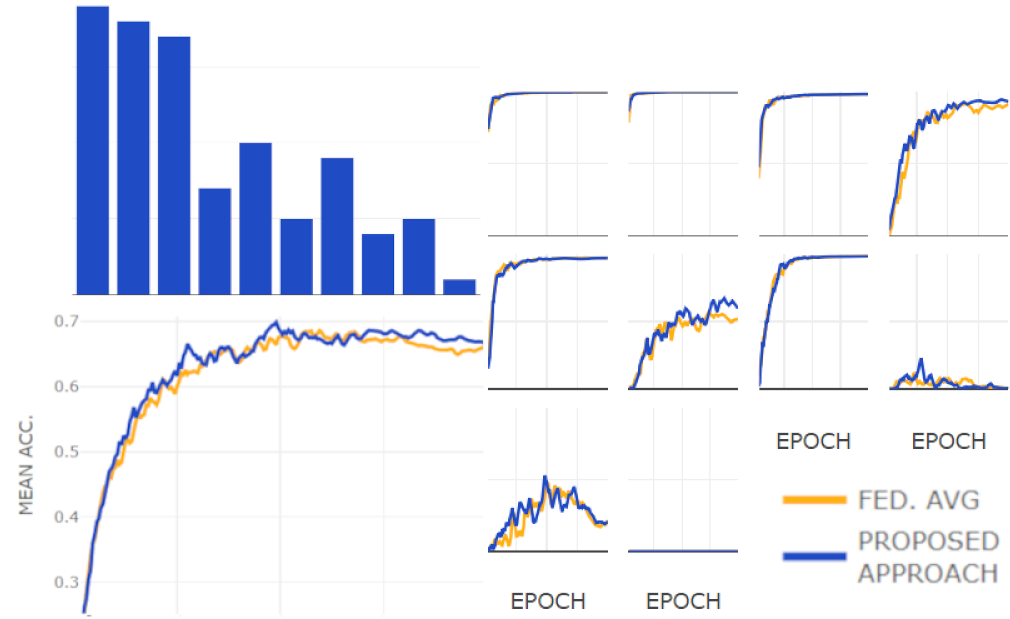}
    \caption{Results of the second MNIST label skew experiment.}
    \label{pop_mnist_label_l}
\end{figure}

The results of the second MNIST experiment, in which the distribution of labels was shifted towards labels 0, 1, and 2, are shown in Figure~\ref{pop_mnist_label_l}. 

The difference in total IID test accuracy between federated pipeline without modifications and the AdFL approach is 1.5-2\%. As previously, two underrepresented labels ``performed better'': the accuracy improvement for the 3rd label is around 5 to 9\%, and the respective improvement for the 5th label is in range from 10 to 14\%.

\begin{figure} [htbp]
    \centering
    \includegraphics[scale=0.5]{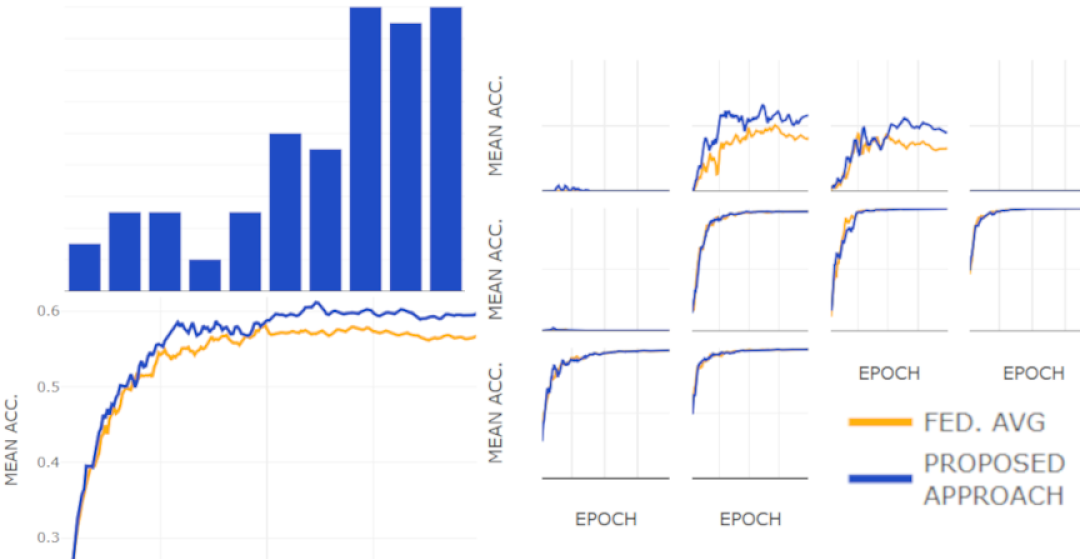}
    \caption{Results of the third MNIST label skew experiment.}
    \label{pop_mnist_label_r}
\end{figure}

The final MNIST experiment featured label distribution being skewed towards labels 7, 8, and 9, with the rest being less populated on local devices (Figure~\ref{pop_mnist_label_r}).
As previously, total IID accuracy improvement was around 3\% over baseline FL, and the source of this improvement came from the 1st label, with accuracy increase in range of 9 to 12\%, and the 2nd label, with gain 11-13\%. 

In summary, for MNIST experiments, the AdFL accuracy improvement came from mildly underrepresented labels (15-30\% of local clients have the label inside the local dataset) with approximately similar performance for other labels, as compared to the baseline FL method. 

\begin{figure} [htbp]
    \centering
    \includegraphics[scale=0.45]{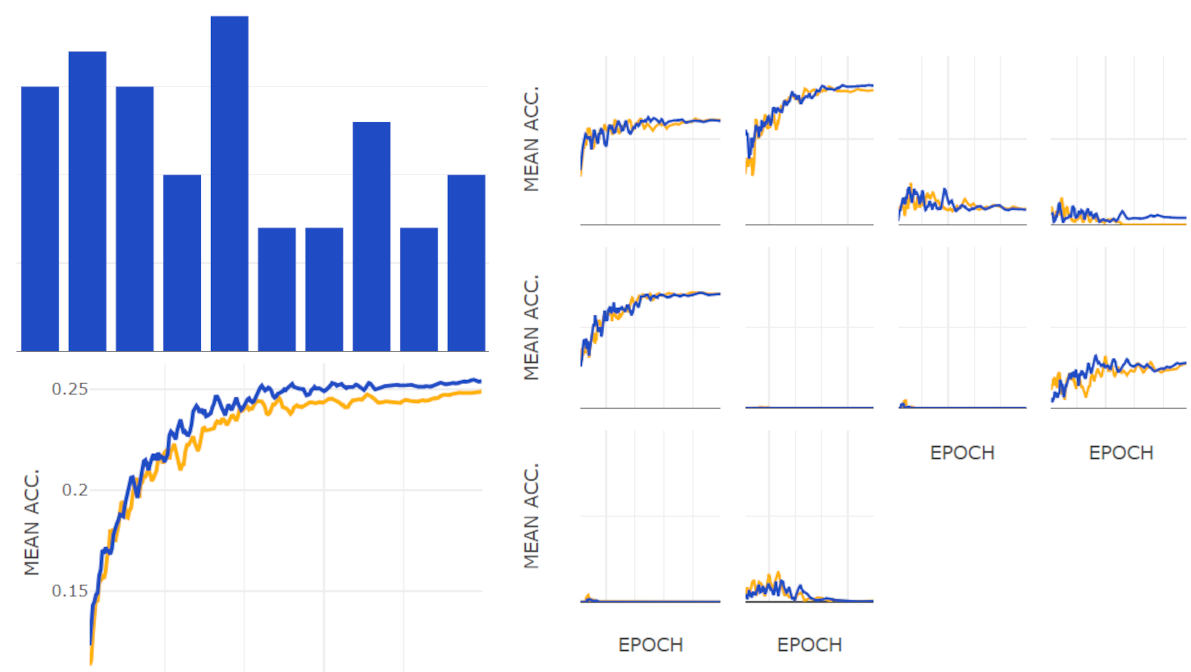}
    \caption{CIFAR-10 label skew experiment.}
    \label{pop_cifar_label_l}
\end{figure}

A similar behaviour can be spotted for the CIFAR-10 dataset. Figure~\ref{pop_cifar_label_l} shows the results of one of the conducted experiments. Label distribution in the experiment is not as severe as in the MNIST experiments -- underrepresented labels are presented in less than 30\% of local clients.

The difference in total IID accuracy between the two compared algorithms is around 1\%, with two labels being predicted better by AdFL: the 1st label improvement is about 2\%, and the 3rd label improvement equals 3.5-4\%.

\section{Concluding remarks}

The initial experiments with the Adversarial Federated Learning (AdFL) algorithm, proposed in this paper, show accuracy improvement on two standard benchmark sets with non-IID label distribution. The accuracy on individual labels raises from 2 to 14\% depending on the specific distribution scheme. Obviously, more experiments and a closer look at their results are required. 

The main areas of future research include (1) verification of the AdFL efficacy on more complex datasets. As the experiments presented in this work were performed with a relatively small number of clients (30), (2) more populated experiments with more nodes and similar non-IID data partitions are also needed to confirm the initial conclusions. Using (3) new datasets will help to understand if the achieved results are data-independent, find cases where adversarial inputs fail to transfer from one client to another and explore new behavior patterns of the algorithm. 

\subsubsection{Acknowledgements} 
Research funded in part by the Centre for Priority Research Area Artificial Intelligence and Robotics of Warsaw University of Technology within the Excellence Initiative: Research University (IDUB) programme.

\bibliography{bibliography}

\end{document}